\documentclass[letterpaper, 10 pt, conference]{ieeeconf}  

\IEEEoverridecommandlockouts                              

\usepackage{graphics} 
\usepackage{graphicx}
\usepackage{booktabs}
\usepackage{multirow}
\usepackage[normalem]{ulem}
\useunder{\uline}{\ul}{}
\usepackage{paralist}
\usepackage{anyfontsize}
\usepackage{verbatim}
\usepackage{cite}
\usepackage{flushend}
\usepackage{amsmath} 
\usepackage{caption}

\let\eqref\undefined
\newcommand{\figref}[1]{Fig.~\ref{fig:#1}}

\newcommand{\secref}[1]{Section~\ref{sec:#1}}
\newcommand{\tabref}[1]{Table~\ref{tab:#1}}
\newcommand{\eqref}[1]{Eq.~\ref{eq:#1}}

\newcommand{\figlabel}[1]{\label{fig:#1}}

\newcommand{\seclabel}[1]{\label{sec:#1}}
\newcommand{\tablabel}[1]{\label{tab:#1}}

\newcommand{\ApproachName}{\textsc{vi-ikd}}

\usepackage{xcolor}

\usepackage{fancyhdr}

\definecolor{green}{RGB}{11,155,13}

\title{\LARGE \bf
VI-IKD: High-Speed Accurate Off-Road Navigation using Learned Visual-Inertial Inverse Kinodynamics
}

\author{Haresh Karnan$^{1*}$, Kavan Singh Sikand$^{2*}$, Pranav Atreya$^{2}$, Sadegh Rabiee$^{2}$, \\Xuesu Xiao$^{2}$, Garrett Warnell$^{2,4}$, Peter Stone$^{2,3}$, and Joydeep Biswas$^{2}$
\thanks{$^{1}$The University of Texas at Austin, Department of Mechanical Engineering {\tt\small haresh.miriyala@utexas.edu}}%
\thanks{$^{2}$The University of Texas at Austin, Department of Computer Science, {\tt\small \{kvsikand, pranavatreya\}@utexas.edu, \{srabiee, xiao, joydeepb, pstone\}@cs.utexas.edu}}%
\thanks{$^{3}$ Sony, AI}%
\thanks{$^{4}$ Computational and Information Sciences Directorate, Army Research Laboratory {\tt\small garrett.a.warnell.civ@army.mil}}%
\thanks{*Equal Contribution}
}

\begin{document}

\maketitle
\thispagestyle{fancy}
\pagestyle{empty}

\begin{abstract}

One of the key challenges in high-speed off-road navigation on ground vehicles is that the kinodynamics of the vehicle-terrain interaction can differ dramatically depending on the terrain.  Previous approaches to addressing this challenge have considered learning an inverse kinodynamics (\textsc{ikd}) model, conditioned on inertial information of the vehicle to sense the kinodynamic interactions. In this paper, we hypothesize that to enable accurate high-speed off-road navigation using a learned \textsc{ikd} model, in addition to inertial information from the \emph{past}, one must also anticipate the kinodynamic interactions of the vehicle with the terrain in the \emph{future}. To this end, we introduce Visual-Inertial Inverse Kinodynamics (\ApproachName{}), a novel learning based \textsc{ikd} model that is conditioned on visual information from a terrain patch ahead of the robot in addition to past inertial information, enabling it to anticipate kinodynamic interactions in the future. We validate the effectiveness of \textsc{vi-ikd} in accurate high-speed off-road navigation experimentally on a scale 1/5 UT-AlphaTruck off-road autonomous vehicle in both indoor and outdoor environments and show that compared to other state-of-the-art approaches, \ApproachName{} enables more accurate and robust off-road navigation on a variety of different terrains at speeds of up to $3.5 m/s$.

\end{abstract}

\section{INTRODUCTION}

Constraining wheeled mobile robot navigation to structured environments and low speeds allows roboticists to use simplified assumptions about the robot's dynamics. Most state-of-the-art classical autonomous navigation systems \cite{joydeep_graphnav, rosmovebase} incorporate motion planners that model a complex kinodynamic system such as a wheeled mobile robot using simplified kinematic models, often ignoring dynamic effects like slippage and wheel suspension. In addition to kinodynamic effects, delays caused by actuation latency inherent in the vehicle's hardware are often ignored. While ignoring such effects at low speeds may be acceptable, the combination of actuation latency coupled with kinodynamic responses due to vehicle-terrain interaction can have a magnified effect on the state of a vehicle when travelling at high speeds, and can be catastrophic (e.g., cause collisions) if not accounted for by the controller. 

While accurate mathematical modelling of such effects is difficult \cite{josiahsim2real, garat, sgat}, recent learning-based approaches to robot navigation have shown promising results in modelling the kinodynamic effects utilizing information from the Inertial Measurement Unit (IMU) to sense the vehicle-terrain interaction. Xiao et al.~\cite{xuesu2021} introduce a learned inverse kinodynamics model (\textsc{ikd}) that enables a ground vehicle to sense the terrain and adaptively navigate at high speeds. This learned \textsc{ikd} model (henceforth called \textsc{imu-ikd}) utilizes inertial sensors on a vehicle to sense the vehicle-terrain interactions and takes a data-driven approach to model the kinodynamic effects experienced by the vehicle on different terrains. However, an inertial sensor is limited in its capability: it can only sense interactions with terrain \emph{after} the vehicle has driven over it. During high speed navigation, latency inherent in the hardware of a vehicle causes actuation commands to be executed at a future world position. Thus, when traversing between terrain types, it is important for the vehicle to \emph{proactively} adjust its controls based on the terrain it is about to encounter in the future, not just the terrain it is currently driving over. A model relying on inertial information alone cannot foresee the kinodynamic response at this future position.
Unlike an inertial sensor, a visual sensor from an egocentric viewpoint enables perception of the world ahead, providing information about the terrain the vehicle will interact with in the future. We therefore hypothesize that in addition to inertial information from the past, conditioning a learned \textsc{ikd} model on the visual information of the terrain ahead will improve the vehicle's capability to accurately navigate at high speeds.

Towards this end, in this paper, we present Visual-Inertial Inverse Kinodynamics (\ApproachName{}), a novel, computationally tractable learning-based approach for incorporating visual information into an inverse kinodynamic model. \ApproachName{} conditions the \textsc{ikd} model on---in addition to inertial information---a visual patch of terrain in the future, by sub-sampling an image captured from a forward-facing camera and extracting only the region where the next actuation command will be executed, considering actuation delays.
Specifically, \ApproachName{} learns a viewpoint-invariant representation of visual terrain patches combined with inertial information captured by an on-board IMU to learn a terrain cognizant \textsc{ikd} model. The resultant \textsc{ikd} model is capable of anticipating the effect of terrain on the robot's dynamics and proactively adapts controls to accurately track planned trajectories on varying types of terrain. 

We evaluate the performance of \ApproachName{} on a scale 1/5 Ackermann-drive vehicle in challenging indoor and outdoor real-world environments with varying types of terrain and demonstrate that it can accurately navigate the robot at high speeds of up to $3.5 m/s$, resulting in improved success rates on the task of reference trajectory following, compared to state-of-the-art approaches.

\section{RELATED WORK}

In this section, we first review related literature on classical methods for wheeled robot navigation in the presence of wheel slippage. We then survey related learning-based approaches for off-road robot navigation. 

\subsection{Physics-Based Kinodynamic Models}
There exists a plethora of research on empirically derived physics-based dynamic and kinodynamic models for wheeled mobile robots that predict the effects of wheel slippage~\cite{seegmiller2013vehicle, rabiee2019friction, tarokh2005kinematics}.
Seegmiller et al.~\cite{seegmiller2013vehicle} propose a parametric kinodynamic model to predict the residual velocity of the robot with respect to the output of a pure kinematic model, given the velocity of the robot and the estimated centrifugal forces. Rabiee et al.~\cite{rabiee2019friction} incorporate an empirical wheel-terrain interaction model into the forward kinematic model of skid-steer robots. 
All of these approaches include a calibration phase that is performed separately for each discrete type of terrain. 
During inference, these methods rely on perception modules to classify the terrain into pre-specified classes using IMU and camera data~\cite{collins2008vibration, wigness2019rugd} in order to switch between different terrain-dependent parameter sets.

\subsection{Error Modelling and Reactive Control}
In off-road unstructured environments, the terrain traversed by the robot cannot be easily delineated into large uniform regions. Instead, there exist frequent transitions between terrain types, e.g. small patches of grass or loose leaves on dirt, such that different robot wheels can be in contact with patches of terrain with significantly different characteristics.
Xiao et al.~\cite{xuesu2021} treat terrain characteristics in a continuous manner and learn an inverse-kinodynamic model that uses a history of IMU data along with the robot's current and desired state to issue control commands. They demonstrate that this approach enables the robot to accurately navigate at high speeds on unstructured terrain without an explicit enumeration of terrain types.
Another line of work that does not require enumeration of terrain types is closed-loop motion control for trajectory following in the presence of slip~\cite{gonzalez2013control, helmick2007terrain, ostafew2016learning}. Koppel et al.~\cite{koppel2016online} learn a statistical model for terrain disturbance using control and visual information. Ostafew et al.~\cite{ostafew2016learning} learn a non-parametric disturbance model online to compensate for slippage that is estimated using visual odometry. These methods are inherently reactive to the sensed changes in terrain characteristics, and therefore only target low-speed navigation applications such as planetary exploration rovers. 
In high-speed navigation, however, the effect of motion control loop delay on trajectory tracking accuracy is significant, as the robot displacement during the period of a control loop is considerable. Sensory information from cameras and LiDAR reveals a great deal about the characteristics of terrain, and can be leveraged to anticipate its effects on the robot's dynamics. While researchers have recently started to incorporate visual information into gait planners for legged-robots~\cite{miki2022learning}, wheeled mobile robot motion planners that use visual information have been mostly limited to end-to-end learning solutions.

\subsection{Learning for Off-Road Navigation}

With the initial success of applying machine learning techniques to mobile robot navigation instead of explicitly modeling the environment and designing complex navigation systems~\cite{xuesusurvey, bojarski2016end, tai2016deep, pfeiffer2017perception, xiao2021toward, xiao2021agile, wang2021agile, tai2017virtual, chiang2019learning, chen2017decentralized, karnan2021voila, scand, manderson2020learning}, roboticists have also applied learning for off-road navigation. 
Pan et al.~\cite{pan2020imitation} propose an end-to-end learning solution that uses camera and odometry data to navigate a high-speed robot on a race track. While such learning-based solutions are appealing for their ability address perception, planning, and control together in a single model, they require large amounts of training data and struggle to generalize to new environments. 
Siva et al.~\cite{siva2021enhancing} enhance ground maneuverability consistency on complex off-road terrain by learning offset behaviors in a self-supervised fashion to compensate for the inconsistency between the actual and expected behaviors without requiring the explicit modeling of various confounding factors. 
Other prior works in the literature have taken a hybrid approach, e.g., learning from visual information for slip-aware robot navigation to estimate the traversal cost of different regions of terrain~\cite{drews2019vision, angelova2007learning, manderson2020learning}.
Angelova et al.~\cite{angelova2007learning} propose a non-parametric method for learning to predict slip on patches of terrain given the appearance and geometric properties perceived by stereo-vision. The resultant information is used to inform the robot to avoid challenging terrain types. Our work, however, seeks to learn to navigate the robot on such challenging terrain as it is unavoidable in unstructured off-road environments.

Our approach is similar to the approach by Xiao et al.~\cite{xuesu2021} in that we learn an inverse kinodynamic model for motion planning without enumerating discrete types of terrain, but we incorporate visual information as well as IMU data in a computationally tractable manner to anticipate the effects of future terrain on the robot's dynamics, making our approach significantly more responsive to variations in terrain characteristics and robust to the effects of actuation latency during high-speed maneuvers. 

\section{METHOD}


In this section we discuss the formulation of the navigation problem and our novel Visual-Inertial Inverse Kinodynamic (\ApproachName{}) approach.

\subsection{Problem Formulation}



The goal of a navigation planner is to incorporate both global and local information to identify a sequence of actions to take a robot from its current state $x_0$ to a target state $x_n$ which it attempts to reach as efficiently and safely as possible. For simplicity of notation, we will treat the robot's traversal through the environment as a sequence of timesteps, which can be arbitrarily small. The planned sequence of states $\{x_0, x_1, ..., x_n\}$ is referred to as the navigation plan. At a given timestep $t \in [0, n)$, the navigation planner is responsible for producing navigation command $u_t$ with the goal of taking the robot from state $x_t$ to $x_{t+1}$.

Given a vehicle state $x_t$, a control input $u_t$, and a world state $w$, the robot's true response upon executing $u_t$ is given by its forward kinodynamic function $f$
\begin{equation}
    x_{t+1} = f (x_t, u_t, w).
\end{equation}

The navigation planner is therefore attempting to find $u_t$ such that: 
\begin{equation}
    u_t = f^{-1}(x_t, x_{t+1}, w)
\end{equation}

In practice, existing navigation planners struggle to accurately model $f^{-1}$, and therefore after executing $u_t$, the resultant robot state  $\hat{x}_{t+1}$ does not match the navigation planner's intended subsequent robot state $x_{t+1}$. There are two primary reasons for this inconsistency: the navigation planner uses a simplified model of the robot's motion response (often considering only the kinematic response), and the world state $w$ is not directly observable, and therefore the planner does not have sufficient information to correctly estimate the effects of $f$.

Recent work has made great strides towards enabling a motion planner to encode complex system dynamics by leveraging deep neural networks, adding a learned inverse kinodynamic module which indirectly captures world state $w$  \cite{xuesu2021}. For example, Xiao et al.~\cite{xuesu2021} introduce the \textsc{imu-ikd} algorithm in which a recent history of the robot's inertial state $S^h_t = \{s_{t-k}, ... s_{t-1}\}$, where $k$ is the length of the history, is used to estimate the world state $w$ for timestep $t$. Specifically, the \textsc{imu-ikd} algorithm \cite{xuesu2021} estimates $f^{-1}$ by learning a function $f^\textrm{IMU}_\theta$ such that:
\begin{equation}
    f^{-1}(x_t, x_{t+1}, w) \approx f^\textrm{IMU}_\theta(x_t, x_{t + 1}, S^h_t)
\end{equation}

Using a history of recent sensor observations relies on the assumption that the current world state $w$ can be predicted from a recent history of inertial observations. However, for an inertial sensor, this may not always be true. For example, a robot driving on bumpy terrain may subsequently encounter smooth terrain, where the inertial response is much different; Even though the smoothness of the terrain ahead where the next actuation command will be executed is a part of the world state that significantly affects the state of the vehicle, an inertial sensor cannot detect this change unless the vehicle drives over the smooth terrain. To address this limitation, in this work, we propose using extereoceptive sensors, specifically RGB images, to help inform the motion planner of the world state $w$ before the vehicle physically interacts with the terrain ahead. A front-facing camera can see parts of the terrain that the robot has not yet encountered, which enables the use of image observations from previous timesteps to help estimate the current world state. We define $\lambda_t$ as the visual terrain information obtained for timestep $t$. Our visual-inertial inverse kinodynamic module therefore attempts to find a function $f^\textrm{VI}_\theta$ which estimates $f^{-1}$ such that:
\begin{equation}
    f^{-1}(x_t, x_{t+1}, w) \approx f^\textrm{VI}_\theta(x_t, x_{t + 1}, S^h_t, \lambda_t)
\end{equation}

\begin{figure}
    \centering
    \includegraphics[width=0.485\textwidth]{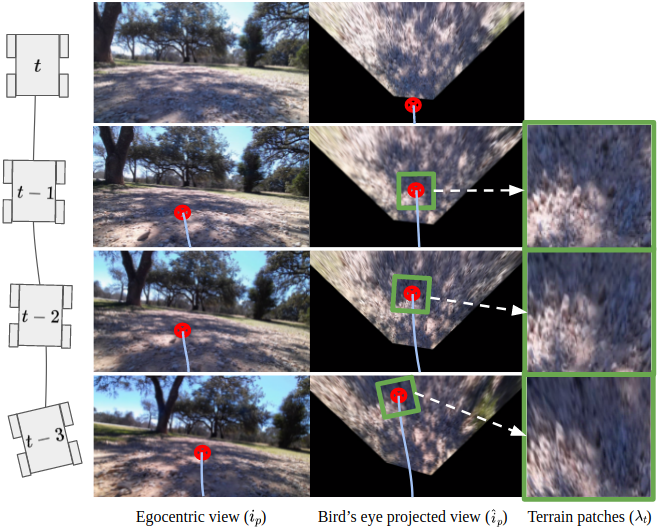}
    \caption{Overview of the Visual Patch Extraction process at time $t$. The robot's next location $\tilde{x}_t$ is estimated (red circle) based on current velocity, and a visible image patch of terrain at the same consistent location $\tilde{x}_t$ is extracted from bird's eye view images from previous timesteps of different viewpoints.}
    \label{fig:patch_extraction}
\end{figure}

\begin{figure}[!th]
    \centering
    \includegraphics[scale=0.32]{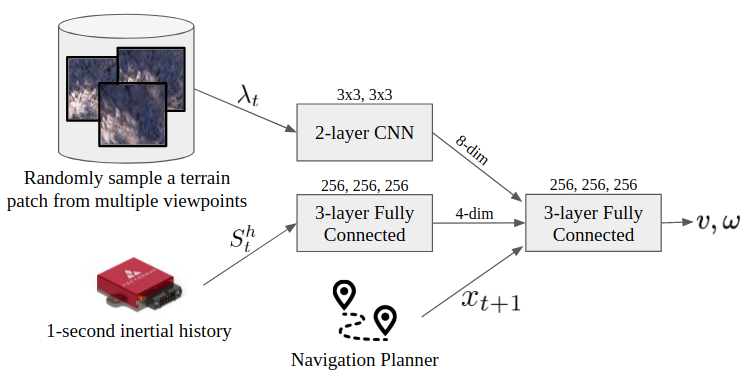}
        \caption{Training setup for the Visual-Inertial Inverse Kinodynamic model. \textsc{vi-ikd} samples one image at random from a set of terrain patches of the same location in ground as viewed from different viewpoints, in every epoch of training. This viewpoint-invariant visual representation combined with inertial information and desired next state is used as input to the \textsc{ikd} model that produces the action commands.}
    \label{fig:visual_ikd_deployment}
\end{figure}

The process for obtaining $\lambda_t$ is given in Sec.~\ref{sec:patch_extraction} and shown in Fig. \ref{fig:patch_extraction}. The process for training $f^\textrm{VI}_\theta$ is given in Sec.~\ref{sec:ikd_learning}. The resulting navigation system is summarized in Fig. \ref{fig:visual_ikd_deployment}.

\subsection{Visual Patch Extraction}
\label{sec:patch_extraction}

When incorporating visual information into the inverse kinodynamic model, it is important to ensure that the visual information used is relevant to the prediction task at hand. Specifically, this should be visual information corresponding to the terrain under the robot at the time a given command is executed, as this is the part of the world state relevant to the kinodynamic response of the robot. As the robot moves through the environment, the front-facing camera captures an egocentric view of the terrain the robot is approaching. The patch of terrain under the robot at any point in time can be extracted from previous camera images with knowledge of the pose information of the robot between frames.

For a particular timestep $t$, a set of captured camera images $I$, a set of IMU measurements $S$, and recent odometry measurements $O$, we seek to find $\lambda_t$, the visual information relevant to the robot's current navigation command. To this end, we define a patch extraction operator $P: \{I, S, O\} \rightarrow \Lambda$ which extracts patches of visual information $\lambda_t \in \Lambda$ of a terrain ahead, from recorded history of observations where the next actuation command will be executed. This operator takes as input a camera image $i_p \in I$ from some timestep $p < t$. This camera image is projected to a birds-eye view (BEV) $\hat{i}_p$ using a homography transform $H$ derived from the static extrinsic camera calibration and $s_p$, the inertial data of the robot at time $p$. We compute the homography transform $H$ in real-time considering the inertial data from the robot due to significant roll-pitch motion experienced during high-speed maneuvers. After this transformation, a fixed distance in BEV projected pixel-space corresponds to a fixed distance in the real world along the ground plane. Once this is done, the robot's real-time recent history of odometry estimates $O$ is used to determine the robot's location relative to the location from which the image was captured. Finally, the robot's current odometry information $o_t \in O$ is used to predict the future location, $\tilde{x}_t$, of the robot in the bird's eye view image plane, where the robot will be at the time when its next issued command will be executed. Note that for a command issued at time $t$ and robot state $x_t$, the command will be executed on the robotic platform at a slightly later state $\tilde{x}_t$ due to actuation latency on a real robot platform. The patch $\lambda$ is then defined as the region of the image around location $\tilde{x}_t$, and is extracted from $\hat{i}_p$. This patch extraction process is shown in Fig. \ref{fig:patch_extraction}. The patches extracted from this process are significantly smaller than a full camera image, enabling \textsc{vi-ikd} to run in real-time.
%

During the training step, \textsc{vi-ikd} uses all observations from different viewpoints of the same consistent location
to learn a viewpoint invariant visual representation of that location. By repeating this procedure for different locations in the world, we ensure that \textsc{vi-ikd} is \emph{viewpoint-invariant} -- that is, it is invariant to observations of the same location irrespective of image variations due to differing observing poses. This procedure also provides robustness to image aberrations and distortion due to artifacts such as motion blur. 

\subsection{Learning Visual-Inertial Inverse Kinodynamics}
To train the \textsc{vi-ikd} module, we collect a set of human demonstrations $D$ in an open environment by teleoperating the vehicle with a joystick.
For each demonstration $d \in D$, we track joystick commands $U$, inertial data $S$, odometry data $O$, and image data $I$, and we record the observed sequence of robot states $X_{obs}$. We then generate training samples of the form $\langle x_{t+1}, x_t, O^h_t, S^h_t, i_t, u_t \rangle$, where $x_{t+1} \in X_{obs}$ is the desired robot state, $x_t \in X_{obs}$ is the preceding state, $S^h_t \subset S$ is the recent inertial history of the robot, $O^h_t \subset O$ is the recent history of odometry measurements, $i_p \in I : p < t$ is a recent camera image, and $u_t \in U$ is the command which transitions the robot from state $x_t$ to $x_{t+1}$. Because we are recording actual observations, these samples encode the true kinodynamic response of the robot, and we know $f^{-1}(x_t, x_{t+1}, w) = u_t$. Given our patch extraction operator $P$, our training loss then seeks to find parameters $\theta$ which minimize 
\begin{equation}
    \arg\min_\theta \sum || u_t - f^\textrm{VI}_\theta ( x_t, x_{t+1}, S^h_t, P(i_p, S^h_t, O^h_t)) ||.
\end{equation}
This learning objective enforces that the \textsc{vi-ikd}-generated control for reaching state $x_{t+1}$ from $x_t$ matches the controls that were actually executed to effect that change.
Note that in this formulation, for each $x_t$, we frequently have multiple different preceding states from which visual information $i_p$ can be extracted, as each traversed patch of terrain may appear in multiple preceding image frames. In these situations, we replicate this for each available choice of $i_p \in I$ from which a patch can be extracted, which helps ensure that regardless of the viewpoint, we learn the same mapping of visual information to predicted command. Regularizing the training process with terrain patches from a consistent location on the ground, but as seen from different viewpoints at different times provides viewpoint invariance in the learned visual representations. In the event where there is no patch information available for a sample, we provide a vector of zeros as the visual representation to the \textsc{ikd} model.

\label{sec:ikd_learning}

\subsection{Implementation Details}

The Visual Inverse Kinodynamic Module $f^\textrm{VI}_\theta$ consists of a visual encoder (2-layer convolutional neural network with a kernel size of 3 and stride of 2), an IMU encoder (3-layer Multi-Layer Perceptron (MLP) with skip connections and hidden layers of size 256), and a final shared 3-layer MLP with skip connections and hidden layers of size 256. To ensure fair comparison, the baseline \textsc{imu-ikd} algorithm uses the same network architecture for the IMU encoder and the \textsc{ikd} network. The network architecture along with the inputs during training time are shown in Fig. \ref{fig:visual_ikd_deployment}. The visual encoder was run off-board at inference time using a GPU-enabled laptop (Nvidia RTX 2060). We regularize the training by randomly sampling a visual terrain patch for a data sample $\langle x_{t+1}, x_t, O^h_t, S^h_t, i_t, u_t \rangle$ from a set of visual terrain patches of the same unique location sub-sampled from observations recorded at previous timesteps. We maintain a buffer of 30 past images to perform patch extraction. The terrain patches are RGB images of fixed size 64-by-64. This patch size was chosen to maximize visual information while ensuring the \textsc{vi-ikd} model can run at 40hz on the GPU with PyTorch and CUDA acceleration. 


\section{Experimental Results}
To evaluate the effectiveness of the Visual-Inertial Inverse Kinodynamic (\textsc{vi-ikd}) model in accurately tracking a trajectory at high speeds, we performed a series of experiments in a controlled indoor environment and an unstructured outdoor environment with different terrains. In this section, we describe the experimental setup followed by the indoor and outdoor experiments.

\begin{figure}[!t]
    \centering
    \includegraphics[scale=0.35]{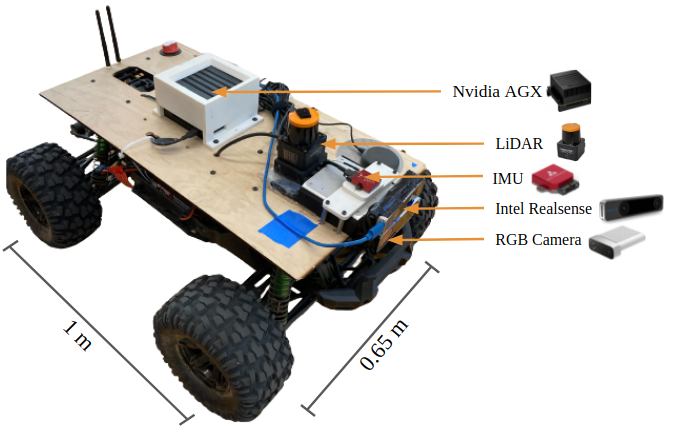}
        \caption{The UT-AlphaTruck scale 1/5th autonomous vehicle and various attached sensors utilized in this work.}
    \figlabel{fig_robot_platform}
\end{figure}

\subsection{Experimental Setup}
We used the same robotic platform for all experiments, pictured in \figref{fig_robot_platform}: the UT-AlphaTruck, a 1/5th scale Ackermann-steer vehicle. The robotic platform is equipped with a Hokuyo planar LiDAR (for obstacle detection and localization), an Intel RealSense T265 tracking camera (for obtaining odometry at 200Hz), a VectorNav VN200 Inertial Measurement unit (for obtaining 6-axis accelerometer and gyroscope measurements at 200Hz), an Azure Kinect camera (for obtaining RGB images at 30Hz), and a Nvidia Xavier AGX (for on-board compute). For the patch extraction procedure, we compute the actuation latency of this hardware to be approximately 0.25 seconds. We do so by subtracting the sensing latency of the RealSense from the sense-act latency (between an issued joystick command and its result as measured by the Intel RealSense).
In our experiments, all methods use a graph-based global planner \cite{joydeep_graphnav} which provides the desired next state $x_{t+1}$ towards a navigation goal. For the indoor experiments, we use Episodic non-Markov Localization (\textsc{e}n\textsc{ml}) \cite{BISWASenml} to track the vehicle's state by fusing LiDAR observations and Intel RealSense's visual-odometry estimates. To collect demonstration data for the training the \textsc{ikd} models, we teleoperate the vehicle using a joystick with $v\in[0, 4] m/s$ and $\omega \in [-1.8, 1.8] rad/s$ for 60 minutes, randomly varying the linear and angular velocities every trajectory. In total, we collect about 32 trajectories containing 73,238 data samples and split them equally into train and test sets. Training the \textsc{vi-ikd} model takes less than 10 minutes on a Nvidia RTX 2060 laptop GPU. We use the same data to train both the \textsc{imu-ikd} and the \textsc{vi-ikd} models.

We compare our method to two alternate approaches:
\begin{itemize}
    \item{\textbf{Baseline}:} The base navigation stack of the Autonomous Mobile Robotics laboratory, which includes a trajectory-rollout based receding horizon local planner that uses a basic kinematic motion model for an Ackermann-drive vehicle \cite{joydeep_graphnav, rabiee2019friction}.
    \item{\textbf{\textsc{imu-ikd}}:} The IMU based \textsc{ikd} model (\textsc{imu-ikd}), introduced by Xiao et al. \cite{xuesu2021}. The \textsc{imu-ikd} model takes as its inputs inertial history $S^h_t$ of the vehicle and a desired next state $x_{t+1}$ to predict a low-level actuation command (forward velocity $v$ and angular velocity $\omega$). 
\end{itemize}
The Visual-Inertial Inverse Kinodynamic (\textbf{\textsc{vi-ikd}}) model utilizes both inertial and visual information from on-board sensors and produces low-level actuation commands based on the desired next state $x_{t+1}$ provided by the global planner. 

\subsection{Indoor Experiments}

\begin{figure}
    \centering
    \includegraphics[width=0.48\textwidth]{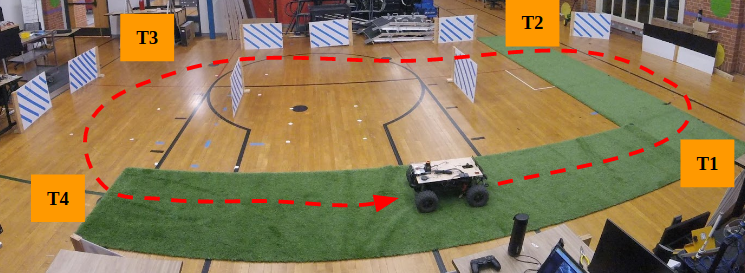}
    \caption{Indoor evaluation environment. Evaluation trajectory is illustrated in red. T1, T2, T3, T4 indicate the four distinct turns in the trajectory. Striped blue rectangular posts are the virtual fixtures used to aid indoor localization using EnML \cite{BISWASenml}.}
    \label{fig:indoor_env}
\end{figure}

\begin{table}
\centering
\caption{Navigation Success Rates in Indoor Environment at $3.2 m/s$}
\begin{tabular}{@{}lcccc@{}}
\toprule
& \textbf{Turn 1} & \textbf{Turn 2} & \textbf{Turn 3} & \textbf{Turn 4} \\
\cmidrule(lr){2-2} \cmidrule(lr){3-3} \cmidrule(lr){4-4} \cmidrule(lr){5-5}
Navigation Controller &
\begin{tabular}[c]{@{}c@{}}Success\\ Count\end{tabular} &
\begin{tabular}[c]{@{}c@{}}Success\\ Count\end{tabular} & \begin{tabular}[c]{@{}c@{}}Success\\ Count\end{tabular} &
\begin{tabular}[c]{@{}c@{}}Success\\ Count\end{tabular} \\ \midrule
Baseline & 0 & 7 & 8 & 5 \\
\textsc{imu-ikd} & 9 & \textbf{10} & \textbf{10} & \textbf{10} \\
\ApproachName{} (Ours) & \textbf{10} & \textbf{10} & \textbf{10} & \textbf{10} \\ \bottomrule
\end{tabular}

\tablabel{indoor_success}
\end{table}

\begin{figure*}
    \centering
    \includegraphics[width=1.0\textwidth]{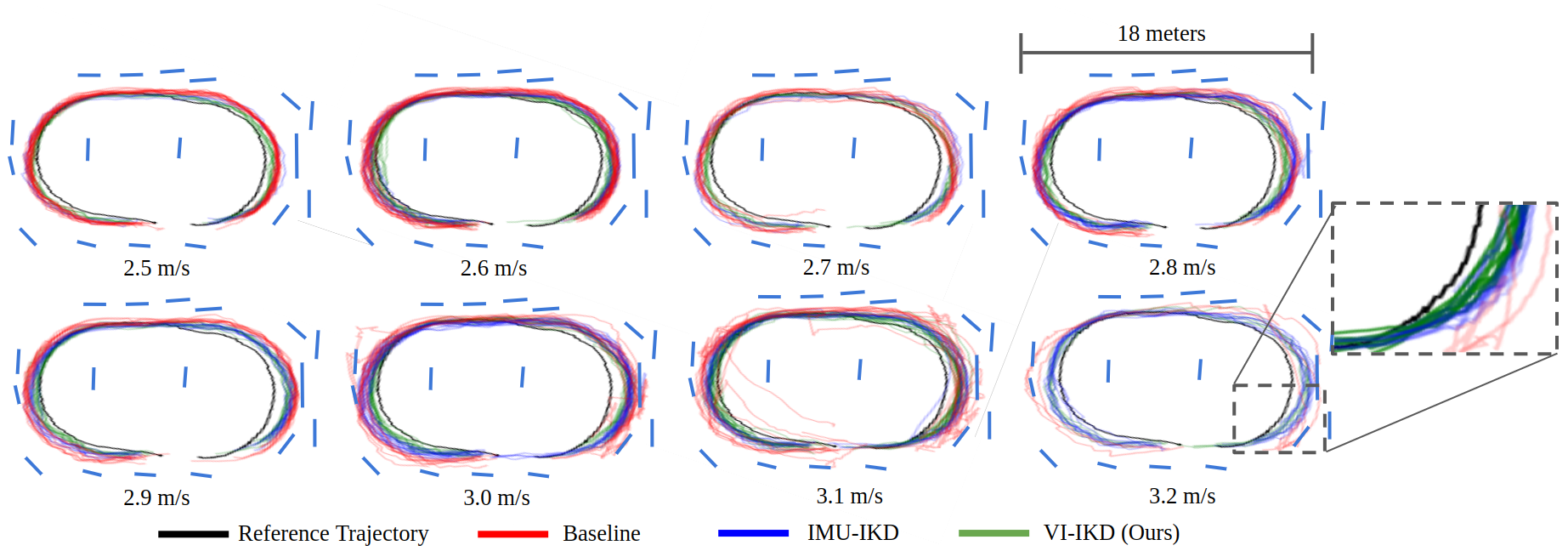}
    \caption{Trajectory traces for the indoor experiments where the vehicle tracks a reference trajectory counter-clockwise at different speeds. The inset shows Turn 1 as executed by the vehicle at $3.2 m/s$ using the three approaches. We see that  \ApproachName{} is able to track the reference trajectory more accurately than \textsc{imu-ikd} \cite{xuesu2021}, confirming our hypothesis. Blue lines along the track show the virtual fixtures used as a map for vehicle state-estimation using \textsc{e}n\textsc{ml} \cite{BISWASenml}.}
    \figlabel{fig_indoor_trajectory_trace}
\end{figure*}

To evaluate the effectiveness of \textsc{vi-ikd}
in accurately and successfully tracking a desired trajectory at high speeds, we set up an indoor course (30 meters long, 15 meters wide) containing two distinct terrain types with different kinodynamic responses at high speeds---wooden floor and green turf---shown in \figref{indoor_env}. The scale 1/5 UT-AlphaTruck vehicle used in these experiments experiences significantly more slip on the wooden floor than on the green turf at high speeds. To aid localization in providing accurate state estimates, we set up virtual fixtures (shown as striped blue posts in Fig. \ref{fig:indoor_env}). To obtain a reference trajectory for navigating this environment, we allowed the robot to autonomously navigate (counter-clockwise) between manually-defined waypoints using the baseline navigation implementation at a slow speed ($0.5 m/s$). At this speed, the impact of dynamics is minimal, and the baseline kinematic motion planner is sufficient for accurate trajectory following. We performed 10 trials for all three navigation systems (baseline, \textsc{imu-ikd} and \textsc{vi-ikd}) at a nominal speed of $2.0 m/s$ and at high speeds ranging from $2.5m/s$ - $3.2 m/s$ in increments of $0.1 m/s$. In total, we perform 270 laps across this loop to evaluate the three approaches. Due to the limited size of the indoor track, the baseline navigation model caused frequent unsafe collisions, preventing us from running experiments at speeds greater than $3.2 m/s$. However, the outdoor experiments presented in \secref{sec_outdoor_experiments} show the potential of \textsc{vi-ikd} to successfully navigate at high speeds of up to $3.5 m/s$.


\begin{figure}
    \centering
    \includegraphics[width=0.4775\textwidth]{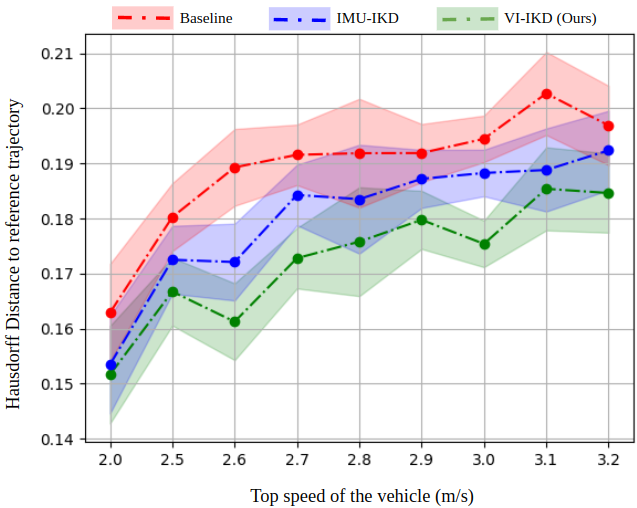}
    \caption{Hausdorff distance (lower is better) between the reference trajectory and trajectories traced by different algorithms at different top speeds of the vehicle in the indoor evaluation environment. We see that \ApproachName{} is the most accurate compared to the receding horizon controller with no \textsc{ikd} model (baseline) and \textsc{imu-ikd} \cite{xuesu2021}.}
    \figlabel{indoor_hausdorff}
\end{figure}

The resultant trajectory traces for each of the navigation systems at various speeds, as well as the reference trajectory, are presented in \figref{fig_indoor_trajectory_trace}. We see that the \textsc{vi-ikd} model introduced in this work is more accurate than the baseline sampling-based local planner and the state-of-the-art \textsc{imu-ikd} model \cite{xuesu2021}. Additionally, we tracked each system's success rate at navigating turns in the environment, where success is any turn that did not result in a collision. We present these success rates when travelling at a maximum speed of $3.2 m/s$ in \tabref{indoor_success}. 
To obtain a quantitative measurement of the accuracy of each navigation system, 
we use an undirected Hausdorff distance, which measures the distance from each point in the trajectory $\Gamma$ to the closest point in the reference trajectory:
\begin{align}
    H(\Gamma_a, \Gamma_b) &= \max (d(\Gamma_a, \Gamma_b), d(\Gamma_b, \Gamma_a)), \\
    d(\Gamma_a, \Gamma_b) &= \max_{a \in \Gamma_a}  \min_{b \in \Gamma_b} ||a - b|| . \nonumber
\end{align}
The results of this numerical evaluation for each navigation system at different navigation speeds is presented in \figref{indoor_hausdorff}. We see that \ApproachName{} is the most accurate compared to the receding horizon controller with no \textsc{ikd} model (baseline) and \textsc{imu-ikd} \cite{xuesu2021}.


\subsection{Outdoor Experiments}
\seclabel{sec_outdoor_experiments}

In addition to the controlled indoor environment, we evaluate \textsc{vi-ikd} in a heterogeneous outdoor environment. We run each navigation system through a fixed set of target waypoints in the environment pictured in \figref{fig_outdoor_env} at a speed of $3.5 m/s$. We provide the reference trajectory to track by manually teleoperating the vehicle around the off-road track. For this trajectory following task outdoors, all algorithms in this experiment use the Intel RealSense's visual-odometry estimates for localization because unlike the controlled indoor experiments, the outdoor track is in off-road, open-ground conditions, unsuitable for accurate LiDAR based localization \cite{BISWASenml}. The outdoor track (50 meters long, 30 meters wide) contains three major turns during which the robot had the potential to slip and deviate from the desired trajectory at high speeds of $3.5 m/s$. Specifically, in Turn 1, the robot makes a u-turn while transitioning from slippery fine sand into grass with increased friction. In Turn 2, the robot transitions between grass, dry leaves, cement and onto pebbles, each producing different kinodynamic responses at high speeds. Finally in Turn 3, the vehicle makes a nearly 180 degree turn on pebbles and enters into a dirt track, which can cause significant slippage at high speeds. Refer to the supplementary video for visual comparisons of the laps performed by the vehicle in this off-road track. The three turns contain significant variance in terrain, requiring an \textsc{ikd} model to anticipate the kinodynamic responses to navigate successfully at high speeds.

\begin{figure}
\centering
    \includegraphics[width=0.49\textwidth]{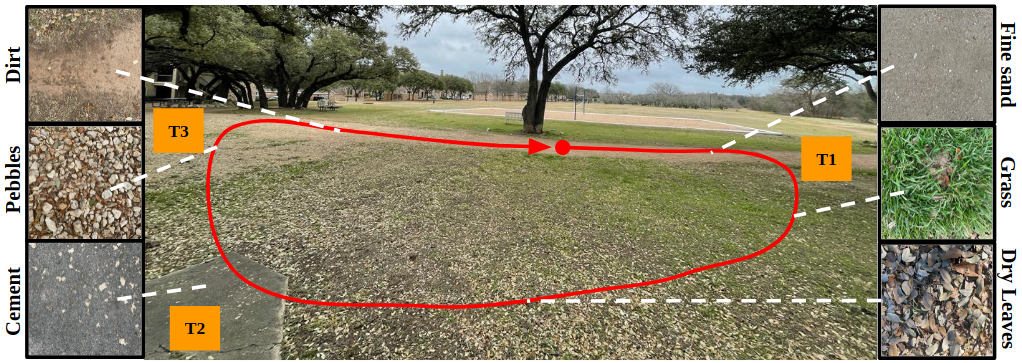}
    \caption{Outdoor Evaluation Environment. Various traversed terrain types are highlighted, and the evaluation trajectory is illustrated in red. T1, T2, and T3 indicate the distinct turns in the trajectory.}
    \figlabel{fig_outdoor_env}
\end{figure}

In our evaluation, each model performs ten laps across this outdoor loop. We mark a turn as unsuccessful if the robot deviates from the desired trajectory beyond the point at which the navigation stack is able to get the robot back on track. At such a failure, we resume trajectory tracking after re-initializing the vehicle in the track at a position after the unsuccessful turn.  In this experiment, we measured the rate at which each navigation system was able to successfully navigate each turn, and present the results of 10 repetitions of the course in \tabref{outdoor_success}. We see that unlike the baseline and \textsc{imu-ikd} model, \textsc{vi-ikd} is able to successfully complete all turns at a high speed of $3.5 m/s$. Although \textsc{imu-ikd} performs better than baseline, the different terrain types present in these turns make it challenging for \textsc{imu-ikd} model to track the reference trajectory without anticipating kinodynamic interactions with the terrain ahead. By anticipating the kinodynamic effects, the \textsc{vi-ikd} model is able to proactively control the vehicle and complete the loops successfully in all 10 trials without any failures. 

\begin{table}
\centering
\caption{Navigation Results in Outdoor Environment at $3.5 m/s$.}
\begin{tabular}{@{}lccc@{}}
\toprule
& \textbf{Turn 1} & \textbf{Turn 2} & \textbf{Turn 3} \\
\cmidrule(lr){2-2} \cmidrule(lr){3-3} \cmidrule(lr){4-4}
Navigation Controller & \begin{tabular}[c]{@{}c@{}}Success\\ Count\end{tabular}
& \begin{tabular}[c]{@{}c@{}}Success\\ Count\end{tabular} & \begin{tabular}[c]{@{}c@{}}Success\\ Count\end{tabular} \\ \midrule
Baseline & 6 & \textbf{10} & 3 \\
\textsc{imu-ikd} & 8 & 7 & 8 \\
\ApproachName{} (Ours) & \textbf{10} & \textbf{10} & \textbf{10} \\ \bottomrule
\end{tabular}

\tablabel{outdoor_success}
\end{table}

\section{CONCLUSION}
In this work, we introduce Visual-Inertial Inverse Kinodynamics (\ApproachName{}), a novel approach for leveraging visual terrain information ahead in addition to inertial information of the past to enhance accuracy in high-speed navigation using a learned \textsc{ikd} model. We hypothesized that utilizing visual information of the terrain helps an \textsc{ikd} model to anticipate kinodynamic effects of the vehicle-terrain interaction and proactively control the vehicle to navigate accurately at high speeds while accounting for actuation delays. Towards this end, the proposed \textsc{vi-ikd} model leverages visual information by learning a viewpoint-invariant representation of the terrain patch ahead, which is used to anticipate kinodynamic responses for the next actuation command executed in the terrain ahead. We validate our hypothesis by comparing \ApproachName{} to state-of-the-art approaches on the task of trajectory following in both indoor and outdoor real-world environments on a scale 1/5 Ackermann-drive vehicle and observe that \ApproachName{} is able to navigate successfully around turns at high speeds of up to $3.5 m/s$ outdoors, and that \ApproachName{} is able to accurately track a reference trajectory at speeds of up to $3.2 m/s$ indoors. 

\section{FUTURE WORK}
There are a few avenues one could pursue to further improve the performance of \textsc{vi-ikd} in future work. First, one could consider a longer control horizon \cite{optimfkd}, rather than the one-step horizon of control we currently use. This would allow the robot to pursue short-term sub-optimal actions to improve long-term utility. Additionally, one could investigate and improve the performance of \textsc{vi-ikd} in unseen terrains, which is essential in off-road conditions where a high-speed vehicle may encounter novel terrains. Finally, one could incorporate additional sensors such as microphones and ground-facing range sensors to further improve the learned terrain representations.

\section*{ACKNOWLEDGEMENT}
\small{
This work has taken place in the Learning Agents Research
Group (LARG) and Autonomous Mobile Robotics Laboratory (AMRL) at UT Austin. LARG research is supported
in part by NSF (CPS-1739964, IIS-1724157, NRI-1925082),
ONR (N00014-18-2243), FLI (RFP2-000), ARO (W911NF19-2-0333), DARPA, Lockheed Martin, GM, and Bosch.
AMRL research is supported in part by NSF (CAREER2046955, IIS-1954778, SHF-2006404), ARO (W911NF-19-2-
0333,W911NF-21-20217), DARPA (HR001120C0031), Amazon, JP Morgan, and
Northrop Grumman Mission Systems. Peter Stone serves as
the Executive Director of Sony AI America and receives financial compensation for this work. The terms of this arrangement
have been reviewed and approved by the University of Texas at
Austin in accordance with its policy on objectivity in research. We would like to thank Corrie Van Sice and Keith Fritz for their assistance with the physical hardware experiments.
}

\bibliographystyle{IEEEtran}
\bibliography{mybib}

\end{document}